\documentclass[10pt,conference]{IEEEtran}
\usepackage{cite}
\usepackage[dvips]{color}
\usepackage{tabu}
\usepackage{comment}
\usepackage{todonotes}
\usepackage{epsf}
\usepackage{epsfig}
\usepackage{times}
\usepackage{epsfig}
\usepackage{graphicx}

\usepackage{mathtools}
\usepackage{mathrsfs}
\usepackage{amssymb,amsmath}
\usepackage{amsmath}
\usepackage{amsfonts}

\usepackage{algorithmicx}  
\usepackage[algo2e]{algorithm2e} 
\usepackage{cleveref}
\usepackage{dsfont}
\usepackage{lettrine} 
\usepackage{epsfig,algorithm,algpseudocode,amsthm,url}
\usepackage{caption}
\usepackage{subcaption}
\usepackage[justification=raggedright]{caption}
\usepackage{subcaption}
\usepackage{bbm}
\usepackage[utf8]{inputenc}
\usepackage{verbatim}
\usepackage{setspace}	
\usepackage[T1]{fontenc}
\usepackage{textcomp}
\usepackage{tabularx}
\usepackage{xpatch}
\DeclareMathOperator*{\argmax}{argmax}

\usepackage{stackengine}
\usepackage[nocomma]{optidef}
\usepackage{commath}

\usepackage[left=1.62cm,right=1.62cm,top=1.85cm,bottom=2.6cm]{geometry}

\IEEEoverridecommandlockouts 

\begin{document}
\title{\Huge Open RAN LSTM Traffic Prediction and Slice Management using Deep Reinforcement Learning 
\thanks{This material is based upon work supported by the National Science Foundation under Grant Numbers  CNS-2202972, CNS- 2318726, and CNS-2232048.}


}
\author{
	\IEEEauthorblockN{
	Fatemeh Lotfi, Fatemeh Afghah}

	\IEEEauthorblockA{Holcombe Department of Electrical and Computer Engineering, Clemson University, Clemson, SC, USA \\
Emails: flotfi@clemson.edu, fafghah@clemson.edu}
}
\maketitle\vspace{-0.cm}
\begin{abstract}

With emerging applications such as autonomous driving, smart cities, and smart factories, network slicing has become an essential component of 5G and beyond networks as a means of catering to a service-aware network. However, managing different network slices while maintaining quality of services (QoS) is a challenge in a dynamic environment. To address this issue, this paper leverages the heterogeneous experiences of distributed units (DUs) in ORAN systems and introduces a novel approach to ORAN slicing xApp using distributed deep reinforcement learning (DDRL). Additionally, to enhance the decision-making performance of the RL agent, a prediction rApp based on long short-term memory (LSTM) is incorporated to provide additional information from the dynamic environment to the xApp. Simulation results demonstrate significant improvements in network performance, particularly in reducing QoS violations. This emphasizes the importance of using the prediction rApp and distributed actors' information jointly as part of a dynamic xApp.\vspace{-0.cm}
\end{abstract}\vspace{-0.cm}

\section{Introduction} \vspace{0cm}

Diverse 6G applications need varying demands across service requirements, encompassing factors such as throughput, latency, traffic capacity, coverage, user density, reliability, and availability. Meeting these stringent demands necessitates the development of network slicing as a solution to maintain the desired network quality of service (QoS) for  different  heterogeneous sets of requirements in the face of dynamic changes~\cite{3gpp2017study, lotfi2022semantic,chinipardaz2022inter}. 

The new open radio access network (ORAN) architecture has been able to maintain heterogeneous service requirements of emerging new technologies. However, different network slices require careful management to avoid diversity in service level agreements (SLAs). The ORAN disaggregates RAN functions into different units, leveraging machine learning (ML) approaches to improve performance. ORAN disaggregated modules include open central units (OCU), open distributed units (ODU), and open radio units (ORU) which are connected to a RAN’s intelligent controllers (RIC) module to control and manage them using artificial intelligence (AI) and ML approaches~\cite{3gpp2017study,oranslice2020,lotfi2022semantic, rajoli2023triplet, samanipour2023stability,ghadermazi2023microbial,sarlak2023diversity,10119039,samanipour2023automated}. The RIC module consists of two distinct components: the near-real-time RIC and the non-real-time RIC, each designed to operate within different control loops. 

The challenges of network slicing in ORAN systems have been explored in recent studies~\cite{lotfi2022evolutionary, lotfi2023attention, cheng2022reinforcement, abdisarabshali2023synergies}. Leveraging the capabilities of ORAN architecture to enhance network performance through ML approaches~\cite{oranslice2020}, several works have delved into dynamic optimization techniques like reinforcement learning (RL) for improving network slicing in ORAN systems~\cite{lotfi2022evolutionary, lotfi2023attention, cheng2022reinforcement, zhang2022team, zhang2022federated}. Some of these approaches employ multiple RL agents to further boost performance. In the studies by \cite{zhang2022team,zhang2022federated,lotfi2023attention}, team learning, federated learning, and distributed agents were employed to leverage collaboration among various intelligent agents. Additionally, other research has explored hybrid methods to enhance the performance of RL approaches~\cite{lotfi2022evolutionary}. 
However, fluctuations in traffic demand over time are non-stationary and unknown in the next-generation wireless networks. 
This can compromise the QoS of broadcasting services. 
Therefore, the continuous management of different network slices is essential to prevent QoS diversity caused by dynamic traffic demand. While some recent studies have addressed traffic prediction in ORAN systems \cite{thaliath2022predictive, kavehmadavani2023intelligent}, they primarily applied their findings to offline network performance optimization. These offline methods lack real-time monitoring capabilities, which can result in SLA violations. 

To address these challenges,  
this paper proposes a novel approach that combines dynamic RL approaches with an RNN prediction to learn and predict ODUs traffic load pattern and improve ORAN network slicing and scheduling task. 
The LSTM-based traffic load prediction is used as rApp in the non-real-time RIC to provide additional information of dynamic wireless network environment for deep RL-based xApp network slicing optimization. Inspired by~\cite{lotfi2023attention} and by considering the potential within the ORAN architecture to create novel experiences through disaggregated modules, we adopt a distributed DRL (DDRL) approach combining with an LSTM-based rApp network traffic load predictor. In this approach, distributed agents are situated at various ODU locations, allowing them to encounter a wide range of network conditions. 
The main contribution of this paper is the integration of the rApp upcoming traffic load prediction results into the RL agent to enhance network slicing agent decision-making by providing additional information from the dynamic wireless network status. 
\emph{To the best of our knowledge, this is the first work that combines an RNN predictor with distributed DRL method for network slicing in ORANs}.

The rest of this paper is structured as follows. Section \ref{sysmodl} outlines the system model and formulates the ORAN slicing problem. Section \ref{proposed} introduces the LSTM-based distributed DRL method and presents the DDRL algorithm designed to address the optimization problem. Section \ref{simulation} offers simulation results. Section \ref{conclusion} summarizes the conclusions drawn from this study.

\vspace{-0.cm}

\section{System model}\label{sysmodl}
Consider an ORAN network structure with a set $\mathcal{L}$ containing $|\mathcal{L}|$=3  types of slices for downlink transmission. 
In this network, $N$ heterogeneous users in a set $\mathcal{N}$ move across the network and are served by different network slices with specific QoS demands as $Q_l$, $l \in \mathcal{L}$. The ORAN $N_m$ 
distributed units (DUs) are distributed in different locations across the network to manage the user equipment (UEs) service requirement resources. Each slice $l \in \mathcal{L}$ in each DU, serves $N_l$  UE with similar QoS. 
Furthermore, the slices must share $K$ common limited resources to meet the QoS requirements of their assigned UEs. 
To assign a proportional resources to each slice and having a suitable scheduling for each UE, we define a wireless model and formulate the joint slicing and scheduling optimization problem. 
\subsection{Wireless communication model}
The QoS for the slices in $\mathcal{L}$ can be expressed as throughput, capacity, and latency. The attainable $Q_l$ is established by utilizing orthogonal frequency-division multiple access (OFDMA) methods. The data rate transmission 
for user $n$ of slice $l$ can be stated as:
\begin{align}\label{urate}
    c_{n,l} = B \sum_{k=1}^{K} e_{n,k} b_{l,k}\log\Big(1+\frac{p_u d_{n}(t)^{-\eta} |h_{n,k}(t)|^2}{ I_{n,k}(t)+ \sigma^2}\Big),  
\end{align}
where the variables $e_{n,k}$ and $b_{l,k}$ indicate resource block (RB) allocation for users and slices, respectively. $B$ is the RB bandwidth, $K$ is the total available RBs for downlink communications, $p_u$ is the transmit power per RB of the radio units (RU), $d_n(t)$ is the distance between user $n$ and its assigned RU, $\eta$ is the path loss exponent, and $|h_{n,k}(t)|^2$ is the time-varying Rayleigh fading channel gain. $I_{n,k}(t)$ denotes downlink interference from neighboring RUs transmitting over RB $k$, and $\sigma^2$ represents the variance of the noise. 

Moreover, the transmission delay for each UE $n$ of slice $l$, can be expressed as:
\begin{align}\label{udelay}
    D_{n,l} = \frac{\Lambda_{n,l}Z_l}{c_{n,l}},
\end{align}
 where $\Lambda_{n,l}$ represents a downlink traffic for user $n$ of slice $l$ in each time slot $t_s$ and $Z_l$ indicates data packet length of slice $l$ services. 

\subsection{Optimization problem}

The objective of the network is to intelligently allocate resources among different slices in a wireless resource-constrained environment to satisfy their distinct QoS demands, while minimizing service level agreement (SLA) violations across the network. To this end, an objective function is defined as $\mathbbm{P}(\abs{Q_l(\boldsymbol{e},\boldsymbol{b})-\lambda_l} \leq \epsilon_l)$ for all slices $l \in \mathcal{L}$. Here, the goal is to determine an optimal resource allocation policy while considering the limitations on the total required resources.
\begin{subequations} 
\begin{align}\label{opt1}
 \argmax_{\boldsymbol{b},\boldsymbol{e}} & \hspace{0.5cm} 
 \mathbbm{P}(\abs{Q_l(\boldsymbol{e},\boldsymbol{b})-\lambda_l} \leq  \epsilon_l),\\
 \text{s.t.,} 
& \hspace{0.5cm} \forall l \in \mathcal{L},\,\,  \forall n \in \mathcal{N} , \label{opt1_q}\\
& \hspace{0.5cm} c_{n,l} \geq \frac{\Lambda_{n,l} Z_l}{D_{n,l}}, \label{opt_delay}\\
& \hspace{0.5cm}  \sum_{l=1}^{L}\sum_{n=1}^{N_l}\sum_{k=1}^{K} b_{l,k}e_{n,k} \leq K, \label{opt1_Ns}\\
& \hspace{0.5cm} \sum_{l}b_{l,k} \leq 1,\,\,   \label{opt1_e}
\end{align}
\end{subequations}\vspace{-0.cm}
where $\mathbbm{P}$ denotes the probability function. 


Slice $l$ has QoS requirements defined by $\lambda_l$ as desired threshold and $\epsilon_l$ as margin. Constraint \eqref{opt_delay} represents users transmission delay and downlink traffic boundary on users data rate. Constraints \eqref{opt1_Ns} and \eqref{opt1_e} enforce the restriction on resource allocation for both slices and UEs in terms of RB availability. Furthermore, the binary variables for resource block allocation make the optimization problem \eqref{opt1} a mixed-integer stochastic optimization which is proven to be an NP-hard problem. To solve the problem we model \eqref{opt1} as a Markov decision process (MDP). However, due to the UEs' movement and the heterogeneity of services, future network traffic load for each DU, i.e ($N_l,\Lambda_{n,l}$), which is required to be known to solve the MDP, is unknown and non-stationary. To address these challenges, we consider a rApp to predict the network unknown traffic load per each DU by training an LSTM model. 
We further define an xApp to address the MDP network slicing problem through a DRL-based approach. 
\vspace{-0.cm} 

\section{Proposed Solution}\label{proposed} \vspace{0cm}
As indicated in \eqref{opt1}, the optimal solution of the optimization problem depends on unknown parameters $N_l$ and $\Lambda_{n,l}$ in each DU. However, obtaining this information requires prior knowledge of the actual traffic of services stored at the data collector in the service management and orchestration (SMO). Consequently, accurate traffic prediction will influence the optimization results. 
To address the network unknown traffic load, i.e. $(N_l,\Lambda_{n,l})$, in a dynamic heterogeneous network, we consider an LSTM recurrent neural network (RNN). The LSTM responsibility is to predict the mentioned parameters by being trained as an rApp at the non-real-time RIC module of the ORAN architecture. 

\subsection{LSTM for traffic prediction rApp}
Due to LSTM capability to retain and selectively update information in its memory cell using the forget ($f_t$), input ($i_t$), and output ($o_t$) gates at time step $t$, LSTM can effectively predict long time series data. The forget gate, $f_t$, allows the LSTM to forget irrelevant past information, while the input gate, $i_t$, allows it to update the memory cell with new relevant information. The output gate, $o_t$, controls the flow of information to the next time step. This gating mechanism prevents the vanishing gradient problem and enables the LSTM to capture long-term dependencies in the data. 


\begin{figure}[t!]
  \centering
    \includegraphics[width=1\columnwidth]{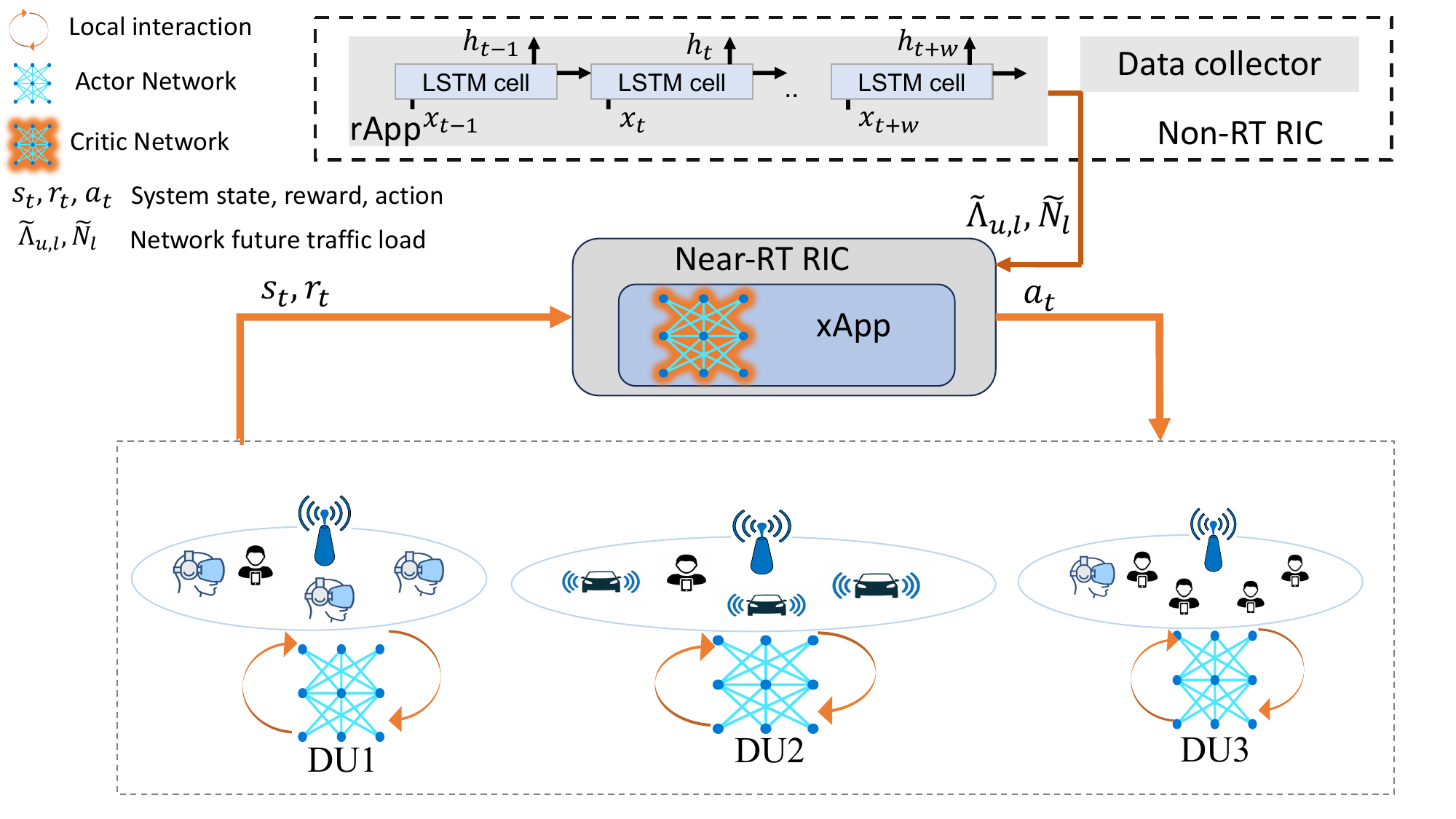}\vspace{-0cm}
    \caption{\small {Implementing an LSTM traffic load prediction over ORAN network structure.} 
    }\vspace{-0cm}
    \label{sys_mdl}
\end{figure}




According to Fig. \ref{sys_mdl}, the sequential connection of LSTM cells in rApp enables the model to capture temporal dependencies and make predictions for time series data. 
In our scenario, the LSTM model plays a vital role in predicting network traffic load, denoted as $N_l$ and $\Lambda_{n,l}$. To achieve this, the LSTM model relies on offline time-series data extracted from the network, which is readily available in the data collector of the non-real-time RIC. 
Fig. \ref{sys_mdl} shows the rApp traffic prediction as an LSTM model in non-real-time RIC in an ORAN architecture.  
The prediction results are subsequently inserted into the xApp module within the near-real-time RIC, enabling the provision of more precise information regarding the network's unknown parameters. 

\subsection{DRL for autonomous slicing xApp}

Consider an intelligent xApp agent, that operates within the near-real-time RIC module, responsible for decision-making in the ORAN slicing environment, as shown in Fig. \ref{sys_mdl}. To this end, the predictive information from the rApp is transmitted to the intelligent xApp agent, to manage ORAN network slicing problem. 
The optimization problem of xApp can be modeled as a Markov decision process (MDP) consisting of tuples $\langle \mathcal{S},\mathcal{A}, T,\gamma,r \rangle$. In this context, $\mathcal{S}$, $\mathcal{A}$, and $T$ refer to the state space, action space, and probability of transition from the current state to the next state (represented by $T=P(s_{t+1}|s_t)$), respectively. Additionally, $\gamma$ and $r$ shows a discount factor and reward function. The tuples for MDP are defined as: 
\subsubsection{State} At every time step, the state of ORAN is represented by $s_t \in \mathcal{S}$, which includes the QoS capability of each slice ($Q_l$), network traffic load ($N_l,\Lambda_{l}$), and the previous resource allocation action taken ($a_{t-1}$). Accordingly, the observation of the intelligent agent at time $t$ is given by $s_t = \{Q_l,N_l,\Lambda_l,a_{t-1}\mid \forall l\in \mathcal{L}\}$. 
\subsubsection{Action} At every time step, the amount of resources required for ORAN slices and UEs is captured by the vector $a_t \in \mathcal{A}$. This vector determines the number of resources that are necessary to meet the demands of the system. Consequently, the intelligent agent employs its policy to select the appropriate action, which is represented as $a_t = \{\boldsymbol{e},\boldsymbol{b}\}$. 
\subsubsection{Reward} The value of $r_t$ is determined by the summation of the SLA violation in each slice $l$. This value is influenced by the incoming traffic for each slice and the radio conditions of the UEs that are connected to the network. The desired reward value is defined as 
   \begin{equation}
        r_t = \sum_{l=1}^{\abs{\mathcal{L}}}\Big( \arctan(\abs{Q_l(\boldsymbol{e},\boldsymbol{b})-\lambda_l}) \Big)^{-1}.
    \end{equation}
The primary aim of every DRL agent is to find an optimal policy $\pi^*(a_t|s_t;\theta_p)$, which functions as a mapping from the state space to the action space. This policy is designed to maximize the expected average discounted reward $\mathbb{E}{\pi}[R(t)]$, where $R(t) = \sum_{i=0}^{\infty}\gamma^i r_{i,t}$.

In our context, due to diverse QoS for each slice, the inverse tangent function is employed to map values to a limited range.  
By taking advantage of the benefits offered by ORAN, employing multiple reinforcement learning (RL) agents situated in distributed units (DUs) throughout the network can improve the training process by utilizing the collective knowledge and experience of all the disaggregated modules. Therefore, in the xApp, we employ a single critic network, while multiple actor networks (totaling $N_m$) are deployed in various DUs throughout the network. Additionally, recognizing the effectiveness of off-policy techniques in using a diverse range of experiences for enough exploration, considering the continuous nature of our state space, we have chosen the Soft Actor-Critic (SAC) algorithm as the most suitable solution. 
The entropy regularization term in SAC algorithm prevents sub-optimal policy solutions and makes this algorithm effective in continuous environments~\cite{haarnoja2018soft}. In this algorithm,  the policy network, represented by the parameter vector $\boldsymbol{\theta}_p$, is updated using the following gradient. 
\begin{align}\label{pupdate} 
    &\nabla_{\theta_p}J(\pi_\theta) = \\
    & \mathbb{E}_{\kappa,\pi}\big[ \nabla_{\theta_p} \log(\pi_{\theta_p}(a|s)) \big(-\beta \log(\pi_{\theta_p}(a|s)) \nonumber+ Q(s,a;\theta_v)\big)\big],
\end{align}
where the symbol $\kappa$ signifies the random sample transitions employed during the parameter updating process, and $\beta$ is a temperature parameter that determines the balance between maximizing entropy and reward. 
In addition, the value network represented by the parameter vector $\boldsymbol{\theta}_v$ will be updated through the minimization of the following loss function, as follows:
\begin{align}\label{vupdate} 
    \min_{\theta_v} \mathbb{E}_{\kappa,\pi} \bigg(y_i-Q_{\pi_i}(s_{i},a_{i};\theta_v)\bigg)^2, 
\end{align}
where $y_i = r_i + \gamma Q_{\pi_{i+1}}(s_{i+1},a_{i+1};\theta_v)-\beta \log(\pi_{\theta_p}(a|s)$.

\begin{algorithm}[t!]
\SetAlgoLined
\textbf{Input}: $N_t$,\,\,$N_m$,\,\,$N_e$\,\,$\theta_{p,i},\forall i \in [0,N_m]$,\,\,$\theta_{f}$,\,\,$\theta_v$.    \\
\For{iteration $t=1:N_t$}{
\For{actor $i=1:N_m$}{
\For{evaluation $e = 1 : N_e$}{
$r_i = \text{evaluate}(\pi_{p,i})$.\\
$\mathcal{B}\gets \langle s_t,a_t,s_{t+1},r_t \rangle $.
}
}
Calculate the upcoming traffic information of the network as $N_l, \Lambda_l, \forall l \in [0,N_m]$. \\
$\mathcal{B}\gets \langle N_l, \Lambda_l \rangle $.\\
Update $\theta_p$ and $\theta_v$ using  \eqref{pupdate} and \eqref{vupdate}.\\
 \If{$\theta_p$ by \eqref{pupdate} is converged}{
 Break.
 }
}
\textbf{Output}: $\theta_{p,i},\forall i \in [0,N_m]$,\,\,$\theta_v$. \\
\caption{The DDRL algorithm}\vspace{-0.cm}
\label{alg1}
\end{algorithm}\vspace{-0.cm}

\subsection{Proposed Algorithm}



The proposed approach combines RL with RNN traffic prediction. This strategy incorporates an LSTM-based prediction rApp within the non-real-time RIC module of the ORAN system. Its purpose is to provide previously unknown traffic information as supplementary data for RL-driven decision-making in the xApp module. Consequently, the LSTM predictor can realize the network's UE patterns, and the xApp's RL agent can leverage historical network data to achieve a more holistic understanding of the environment.

Algorithm \ref{alg1} provides an overview of the proposed DDRL strategy for addressing the optimization problem in \eqref{opt1}-\eqref{opt1_e}. The algorithm takes several input variables into account, including the number of iterations denoted as $N_t$, the number of distributed actors indicated as $N_m$, the number of evaluations labeled as $N_e$, and the initial random initialization of weights for the distributed actors' network and critic network, represented as $\theta_{p,i}$ and $\theta_v$. Additionally, the LSTM prediction network weights in rApp, denoted as $\theta_f$, which have been pre-trained using past network states. During each time step of the algorithm's training process, the predictor network computes the upcoming step's traffic information. This information is subsequently provided to the RL agent located in xApp, serving as part of the agent's environmental state input for its RL algorithm. Next, by considering $N_l$ as a component of the RL state, and $\Lambda_l$ as a necessary parameter in the computation of $Q_l$ according to \eqref{udelay}, the DRL parameters are updated by using \eqref{pupdate} and \eqref{vupdate}. The algorithm concludes upon the convergence of the policy network of the distributed DRL actors or after reaching a maximum of $N_t$ iterations. \vspace{-0.cm}

\section{Simulation Results}\label{simulation}\vspace{-0cm}
To evaluate our approach, we investigate an ORAN architecture with three slices (i.e., eMBB, MTC, URLLC). We consider a scenario with $N_m = 7$ distributed DUs, where one DU is placed at the central point of the cell, and the remaining six DUs are distributed at the corners of two diamond shapes with common corner in the central point. The available bandwidth is $20$ MHz corresponding to $K = 100$ RBs dynamically assigned to different slices. Additionally, there are $60$ users distributed uniformly across a network with one cell covering a $3000 m \times 2000 m$ area. The users are assigned to different slices based on their service demands, denoted as $N_l =\{20,30,10\}$. The mobility of these users is characterized by dynamic speed and direction. User speed is randomly selected between $10 m/s$ to $20 m/s$, and their direction is randomly chosen from a set of seven directions, $\{\pm \pi/3,\pm \pi/6,\pm \pi/12,0\}$. 
The traffic demand for users is determined by their preassigned services, with traffic averages set at $3$ Mbps, $150$ Kbps, and $750$ Kbps. For simplicity, we assume users operate in one of four traffic modes: ${idle, low, mid, high}$. The dynamic changes in traffic modes occur with a probability of $0.01$ at each time step~\cite{cheng2022reinforcement}. 
These traffic modes influence the RB allocation in the scheduling phase, reflecting how much a user utilizes its allocated bandwidth based on the service demand.  
Utilizing the rApp RNN model with $2$ layers of $50$ LSTM units and an $18$ time steps look-back window, the intelligent agent assesses the traffic load demand for each slice at every time step in the RL approach. This evaluation considers user density in each DU and their corresponding traffic profiles. The LSTM model is trained on a Python-generated dataset of simulated environments to predict network traffic load demands as users' density and traffic profile of all users in each DUs for future time steps. 
A DRL approach with PyTorch's actor-critic method was implemented, utilizing three fully connected layers ($128$, $256$, $256$ neurons) for actor and critic networks, a \textit{tanh} activation function, a learning rate of $10^{-4}$, and the \textit{Adam} optimizer for all models.

\begin{figure}[t!]
  \centering
    \includegraphics[width=7cm]{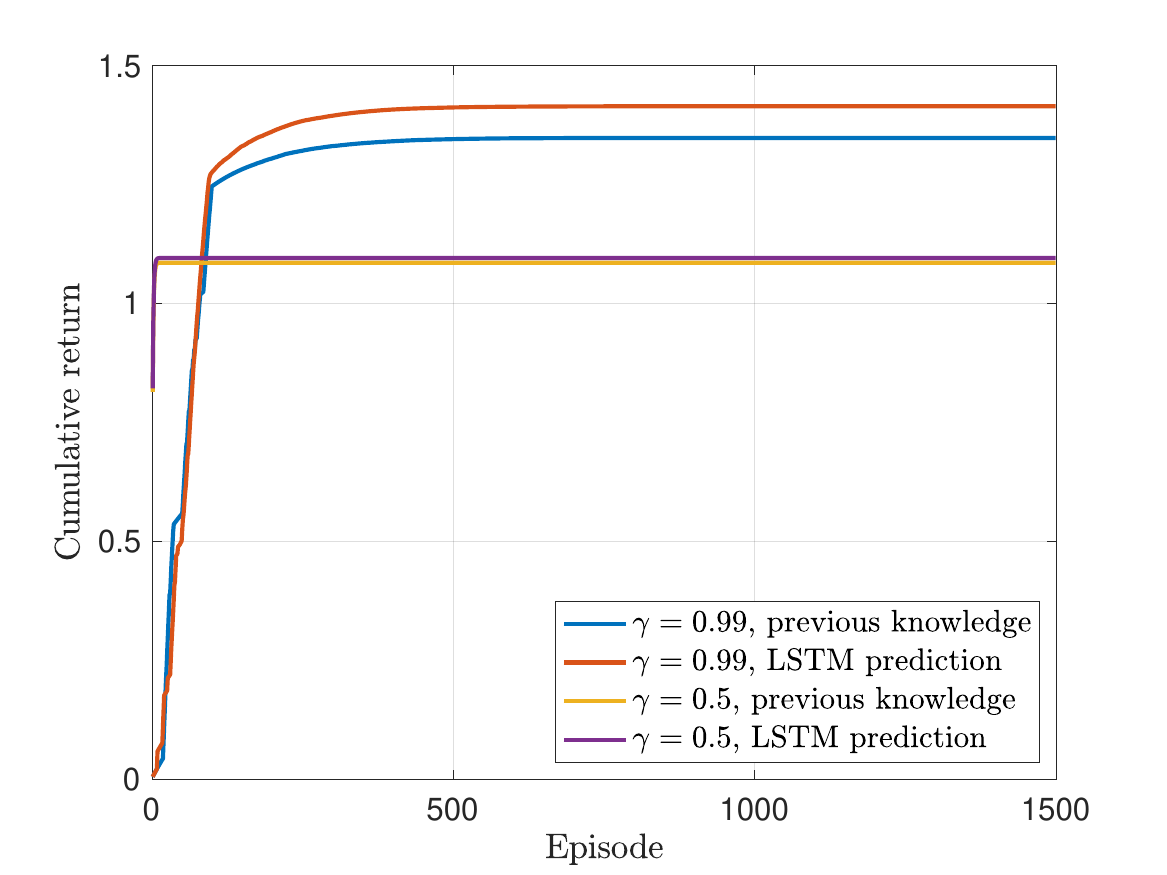}\vspace{-0.cm}
    \caption{\small Convergence analysis and performance comparison of distributed deep RL in different scenarios.}\vspace{-0.cm}
    \label{cum_return}
\end{figure}
Figure \ref{cum_return} shows convergence analysis of the distributed DRL approach in four different scenarios. The cumulative rewards were computed in four scenarios: two scenarios utilized a discount factor of $\gamma=0.99$, emphasizing a long-term perspective in planning RB allocations, while the other two scenarios employed $\gamma=0.5$, placing greater emphasis on current QoS violations. The reported results represent averages over a substantial number of runs. 
Figure \ref{cum_return} displays the results in each episode for comparison. The results reveal that the LSTM prediction based distributed-DRL approach can provide up to a $7.7\%$ greater final return value than distributed-DRL approach without utilizing prediction results, demonstrating the efficacy of using prediction information in the proposed LSTM-based DDRL algorithm over the wireless environment. Figure \ref{cum_return} displays results that indicate the LSTM prediction-based distributed-DRL approach performs better in scenarios where a long-term perspective is crucial.


\begin{figure}[t!]
     \centering
     \begin{subfigure}[a]{0.18\textheight}
         \centering
         \includegraphics[width=\textwidth]
         {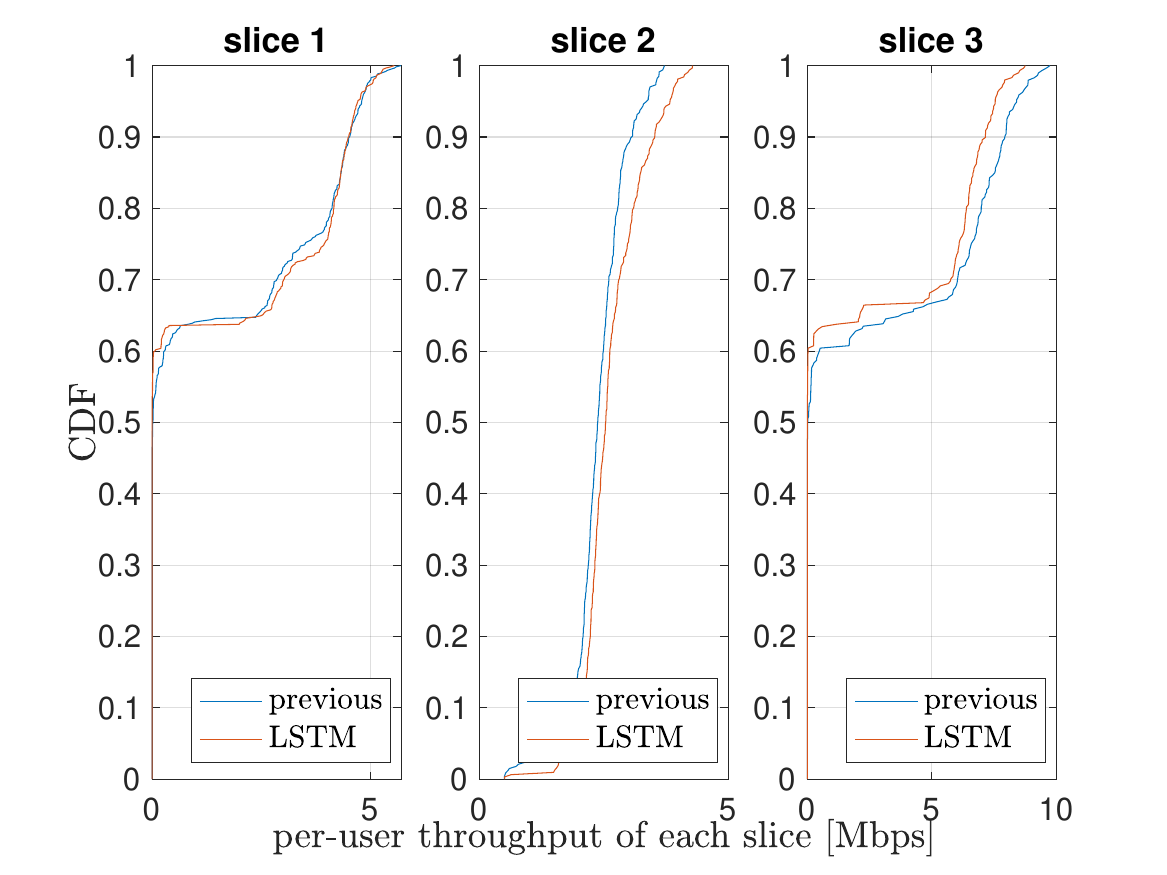}
         \caption{CDF of achieved throughput per users}
         \label{cdf_per}
     \end{subfigure}
     \hfill
     \begin{subfigure}[a]{0.18\textheight}
         \centering
         \includegraphics[width=\textwidth]
         {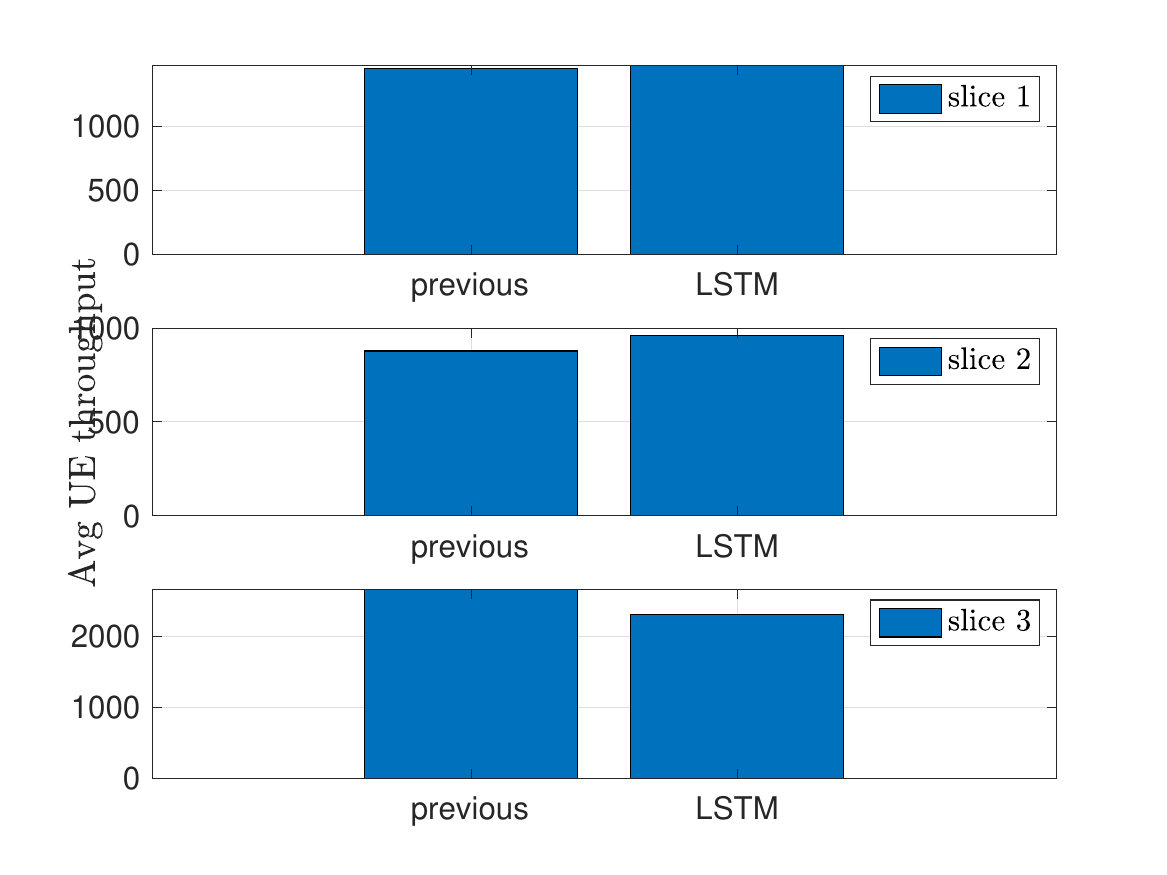}
         \caption{Average of achieved throughput per users}
         \label{avg_per}
     \end{subfigure}
      \caption{achieved throughput per users through distributed deep RL. 
      }\vspace{-0.cm}
        \label{cdf_UE}
\end{figure}

Figure \ref{cdf_UE} displays per-user throughput for each slice in the simulated ORAN environment under two scenarios: with and without utilizing predicted information. Figure \ref{cdf_per} illustrates the cumulative distribution function (CDF) of achieved throughput per user in each slice, while Figure \ref{avg_per} presents the average achieved throughput per user in each slice. The results indicate that, despite the algorithm's main objective being the reduction of SLA violations for users in different slices, there is an observed improvement in their throughput. This implies that while the algorithm primarily focuses on reducing SLA violations and does not explicitly address QoS improvement, it has a positive impact on it.

\begin{figure}[t!]
  \centering
    \includegraphics[width=0.8\columnwidth]
    {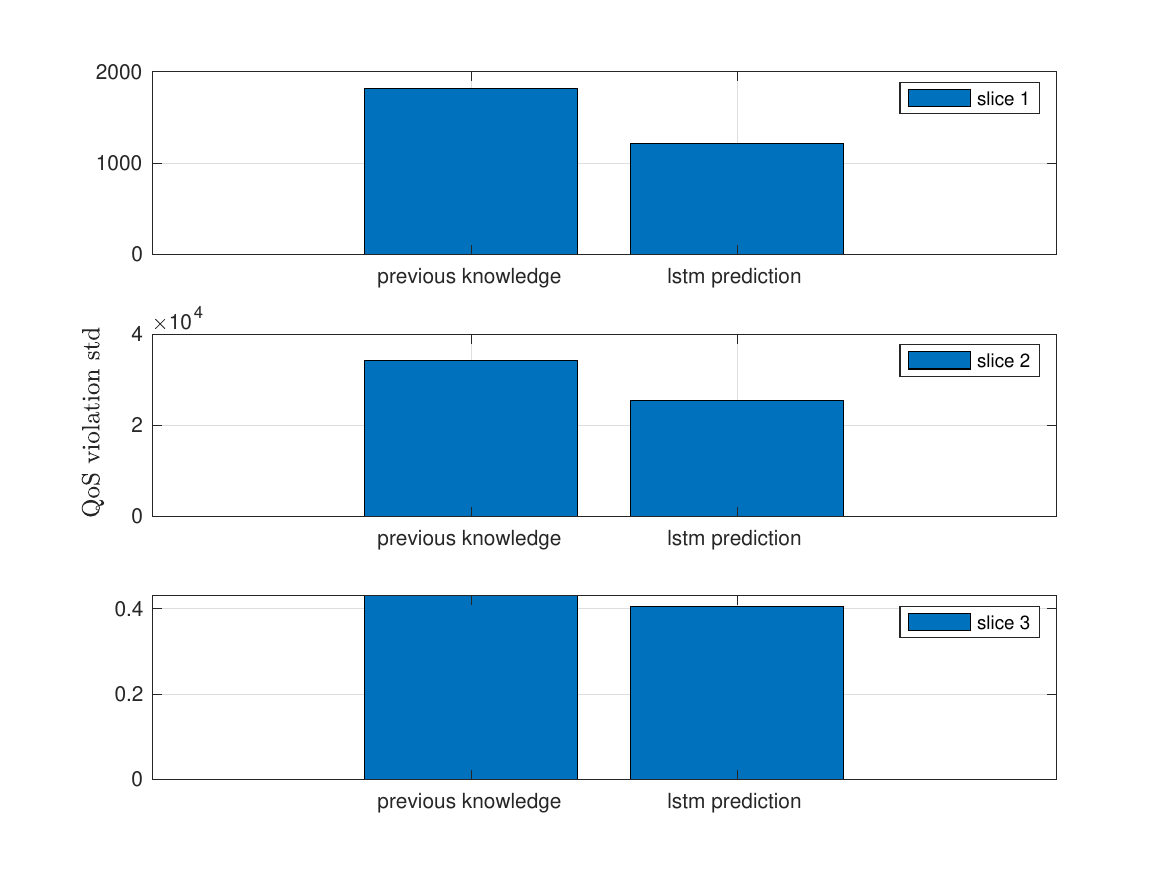}\vspace{-0.cm}
    \caption{\small QoS violation std of each slice for distribution deep RL in different scenarios}\vspace{-0.cm}
    \label{std_q}
\end{figure}

Figure \ref{std_q} compares the standard deviation values (std) of QoS violations for each slice when employing the suggested distributed DRL approach in scenarios with and without prediction information. The primary goal was to minimize QoS violations, and the outcomes demonstrate the efficacy of the proposed LSTM-based DDRL solution in effectively reducing SLA violations.

\section{Conclusion}\label{conclusion}
We introduce a novel framework for ORAN slicing, incorporating a predictive rApp and an optimization slicing xApp modeled as a MDP. Employing a distributed deep RL approach, our model develops network slice management. An LSTM model is employed for predicting network upcoming traffic load located in non-RT RIC as rApp, enhancing decision-making by the distributed DRL actors and a global critic as xApp in near-RT RIC. Simulation results demonstrate the efficacy of utilizing LSTM-based predicted information in ORAN network slice management. 

\vspace{-0.cm}

\def\baselinestretch{0.97}
\bibliographystyle{IEEEbib}
\bibliography{Main}

\begin{thebibliography}{10}

\bibitem{3gpp2017study}
3GPP TR~38.912 version 14.1.0 Release~14,
\newblock ``Study on new radio access technology: Radio access architecture and
  interfaces,''
\newblock {\em Tech. Rep}, , no. 3, 2017.

\bibitem{lotfi2022semantic}
F.~{Lotfi}, O.~{Semiari}, and W.~{Saad},
\newblock ``Semantic-aware collaborative deep reinforcement learning over
  wireless cellular networks,''
\newblock in {\em ICC 2022-IEEE International Conference on Communications}.
  IEEE, 2022, pp. 5256--5261.

\bibitem{chinipardaz2022inter}
Maryam Chinipardaz, Seyed~Majid Noorhosseini, and Ahmad Sarlak,
\newblock ``Inter-cell interference in multi-tier heterogeneous cellular
  networks: modeling and constraints,''
\newblock {\em Telecommunication Systems}, vol. 81, no. 1, pp. 67--81, 2022.

\bibitem{oranslice2020}
O-RAN Working~Group 1,
\newblock ``Study on o-ran slicing-v2.00,''
\newblock {\em O-RAN.WG1.Study-on-O-{RAN}-Slicing-v02.00 Technical
  Specification}, April 2020.

\bibitem{rajoli2023triplet}
H.~{Rajoli}, F.~{Lotfi}, A.~{Atyabi}, and F.~{Afghah},
\newblock ``Triplet loss-less center loss sampling strategies in facial
  expression recognition scenarios,''
\newblock in {\em 2023 57th Annual Conference on Information Sciences and
  Systems (CISS)}. IEEE, 2023, pp. 1--6.

\bibitem{samanipour2023stability}
P.~{Samanipour} and H.~{Poonawala},
\newblock ``Stability analysis and controller synthesis using
  single-hidden-layer relu neural networks,''
\newblock {\em IEEE Transactions on Automatic Control}, 2023.

\bibitem{ghadermazi2023microbial}
P.~{Ghadermazi} and S.~{Chan},
\newblock ``Microbial interactions from a new perspective: Reinforcement
  learning reveals new insights into microbiome evolution,''
\newblock {\em bioRxiv}, pp. 2023--05, 2023.

\bibitem{sarlak2023diversity}
A.~{Sarlak}, X.~{Chen}, R.~{Amin}, and A.~{Razi},
\newblock ``Diversity maximized scheduling in roadside units for traffic
  monitoring applications,''
\newblock {\em arXiv preprint arXiv:2306.16481}, 2023.

\bibitem{10119039}
Ali Owfi, Fatemeh Afghah, and Jonathan Ashdown,
\newblock ``Meta-learning for wireless interference identification,''
\newblock in {\em 2023 IEEE Wireless Communications and Networking Conference
  (WCNC)}, 2023, pp. 1--6.

\bibitem{samanipour2023automated}
Pouya Samanipour and Hasan~A Poonawala,
\newblock ``Automated stability analysis of piecewise affine dynamics using
  vertices,''
\newblock {\em arXiv preprint arXiv:2307.03868}, 2023.

\bibitem{lotfi2022evolutionary}
F.~{Lotfi}, O.~{Semiari}, and F.~{Afghah},
\newblock ``Evolutionary deep reinforcement learning for dynamic slice
  management in {O-RAN},''
\newblock in {\em 2022 IEEE Globecom Workshops (GC Wkshps)}. IEEE, 2022, pp.
  227--232.

\bibitem{lotfi2023attention}
F.~{Lotfi}, F.~{Afghah}, and J.~{Ashdown},
\newblock ``Attention-based open ran slice management using deep reinforcement
  learning,''
\newblock {\em arXiv preprint arXiv:2306.09490}, 2023.

\bibitem{cheng2022reinforcement}
N.~{Cheng}, T.~{Pamuklu}, and M.~{Erol-Kantarci},
\newblock ``Reinforcement learning based resource allocation for network slices
  in o-ran midhaul,''
\newblock {\em arXiv preprint arXiv:2211.07466}, 2022.

\bibitem{abdisarabshali2023synergies}
P.~{Abdisarabshali}, N.~{Accurso}, F.~{Malandra}, W.~{Su}, and
  S.~{Hosseinalipour},
\newblock ``Synergies between federated learning and o-ran: Towards an elastic
  virtualized architecture for multiple distributed machine learning
  services,''
\newblock {\em arXiv preprint arXiv:2305.02109}, 2023.

\bibitem{zhang2022team}
H.~{Zhang}, H.~{Zhou}, and M.~{Erol-Kantarci},
\newblock ``Team learning-based resource allocation for open radio access
  network (o-ran),''
\newblock in {\em ICC 2022-IEEE International Conference on Communications}.
  IEEE, 2022, pp. 4938--4943.

\bibitem{zhang2022federated}
H.~{Zhang}, H.~{Zhou}, and M.~{Erol-Kantarci},
\newblock ``Federated deep reinforcement learning for resource allocation in
  o-ran slicing,''
\newblock in {\em GLOBECOM 2022-2022 IEEE Global Communications Conference}.
  IEEE, 2022, pp. 958--963.

\bibitem{thaliath2022predictive}
J.~{Thaliath}, S.~{Niknam}, S.~{Singh}, R.~{Banerji}, N.~{Saxena},
  H.~{Dhillon}, J.~{Reed}, A.~{Bashir}, A.~{Bhat}, and A.~{Roy},
\newblock ``Predictive closed-loop service automation in o-ran based network
  slicing,''
\newblock {\em IEEE Communications Standards Magazine}, vol. 6, no. 3, pp.
  8--14, 2022.

\bibitem{kavehmadavani2023intelligent}
F.~{Kavehmadavani}, V.~{Nguyen}, T.~{Vu}, and S.~{Chatzinotas},
\newblock ``Intelligent traffic steering in beyond 5g open ran based on lstm
  traffic prediction,''
\newblock {\em IEEE Transactions on Wireless Communications}, 2023.

\bibitem{haarnoja2018soft}
T.~{Haarnoja}, A.~{Zhou}, P.~{Abbeel}, and S.~{Levine},
\newblock ``Soft actor-critic: Off-policy maximum entropy deep reinforcement
  learning with a stochastic actor,''
\newblock in {\em International conference on machine learning}. PMLR, 2018,
  pp. 1861--1870.

\end{thebibliography}
\end{document}